\documentclass{article}
\usepackage{spconf,amsmath,graphicx}
\usepackage{subfigure}
\usepackage{multirow}
\usepackage{array}
\usepackage{color}

\newcommand{\ie}{\textit{i}.\textit{e}. }
\newcommand{\eg}{\textit{e}.\textit{g}., }

\title{ULDor: A Universal Lesion Detector for CT Scans with Pseudo Masks and Hard Negative Example Mining}

 \name{You-Bao Tang$^{\dagger}$ \quad Ke Yan$^{\dagger \star}$ \quad Yu-Xing Tang$^{\dagger \star}$ \quad Jiamin Liu$^{\dagger \star}$ \quad Jing Xiao$^{\ddagger}$ \quad Ronald M. Summers$^{\dagger}$ \thanks{$^{\star}$ indicates equal contribution}}

 \address{$^{\dagger}$ Imaging Biomarkers and Computer-Aided Diagnosis Laboratory, Radiology and Imaging Sciences, \\ National Institutes of Health Clinical Center, Bethesda, MD 20892, USA \\
     $^{\ddagger}$Ping An Insurance Company of China, Shenzhen, 510852, China}

\begin{document}
%
\maketitle

\begin{abstract}
Automatic lesion detection from computed tomography (CT) scans is an important task in medical imaging analysis. It is still very challenging due to similar appearances (\eg intensity and texture) between lesions and other tissues, making it especially difficult to develop a universal lesion detector. Instead of developing a specific-type lesion detector, this work builds a \textit{U}niversal \textit{L}esion \textit{D}etect\textit{or} (ULDor) based on Mask R-CNN, which is able to detect all different kinds of lesions from whole body parts. As a state-of-the-art object detector, Mask R-CNN adds a branch for predicting segmentation masks on each Region of Interest (RoI) to improve the detection performance. However, it is almost impossible to manually annotate a large-scale dataset with pixel-level lesion masks to train the Mask R-CNN for lesion detection. To address this problem, this work constructs a pseudo mask for each lesion region that can be considered as a surrogate of the real mask, based on which the Mask R-CNN is employed for lesion detection. On the other hand, this work proposes a hard negative example mining strategy to reduce the false positives for improving the detection performance. Experimental results on the NIH DeepLesion dataset demonstrate that the ULDor is enhanced using pseudo masks and the proposed hard negative example mining strategy and achieves a sensitivity of 86.21\% with five false positives per image.
\end{abstract}


\section{Introduction}
\label{sec:intro}
Recently, many useful applications in medical image analysis have been proposed, \eg measurement estimation \cite{tang2018semi}, lung segmentation \cite{jin2018ct}, image enhancement \cite{tang2018ct}, lesion detection \cite{yan2018deeplesion} and segmentation \cite{cai2018accurate}, etc. 
Automatic lesion detection from computed tomography (CT) scans plays an important role in computer-aided diagnosis tasks, \eg cancer patient screening, tumor classification and lesion segmentation. By coupling with a lesion detection framework, many existing techniques \cite{tang2018semi,cai2018accurate} can be made fully automatic. Although lesion detection has been extensively studied and many advancements have been achieved, it is still very challenging due to similar appearances (\eg intensity and texture) between lesions and other tissues. Generally, previous works on lesion detection focused on specific lesion types, \eg pulmonary nodules \cite{setio2016pulmonary}. It is more difficult to develop a universal detector to detect all types of lesions from whole body parts. To the best of our knowledge, the first work to collect a large-scale lesion dataset containing different kinds of lesions across whole body parts and develop a universal lesion detector was by Yan et al. \cite{yan2018deeplesion}. 

In \cite{yan2018deeplesion}, Faster R-CNN \cite{ren2017faster} is directly used for lesion detection, where a Region Proposal Network proposes candidate object bounding boxes and the Region of Interest pooling (RoIPool) extracts features from each candidate box and performs classification and bounding-box regression. It doesn't perform the task of segmentation mask prediction in a pixel-to-pixel manner. However,  it has been proved that Mask R-CNN \cite{he2017mask} extending Faster R-CNN can boost the detection performance by adding a branch for predicting segmentation mask on each Region of Interest (RoI). Inspired by this fact, this paper proposes a \textit{U}niversal \textit{L}esion \textit{D}etect\textit{or} (ULDor) based on Mask R-CNN. However, manually annotating a large number of lesion masks is highly tedious and time consuming. To avoid this issue, a pseudo mask is constructed based on the RECIST (Response Evaluation Criteria In Solid Tumors) annotation for each lesion region. With these pseudo masks, the Mask R-CNN model can be trained feasibly and effectively. Also in \cite{yan2018deeplesion}, lots of false positives are produced, which certainly affects the sensitivity performance when the number of false positives is small per image. Therefore, this work proposes a hard negative example mining strategy to reduce the false positives. With the help of these mined negative examples, the Mask R-CNN model can learn more discriminative features to distinguish true lesions from false positives, so as to improve the detection performance.

In summary, the main contributions of this paper are two-fold: (1) A pseudo mask augmented lesion detection method is proposed based on Mask R-CNN, which achieves better detection performance than the one without using pseudo masks. (2) A hard negative example mining strategy is proposed to reduce the false positives and re-rank the orders of true positives forward, which further boosts the detection performance.

\section{The proposed ULDor}

\begin{figure}[t!]
	\begin{center}
		\includegraphics[width=0.99\linewidth]{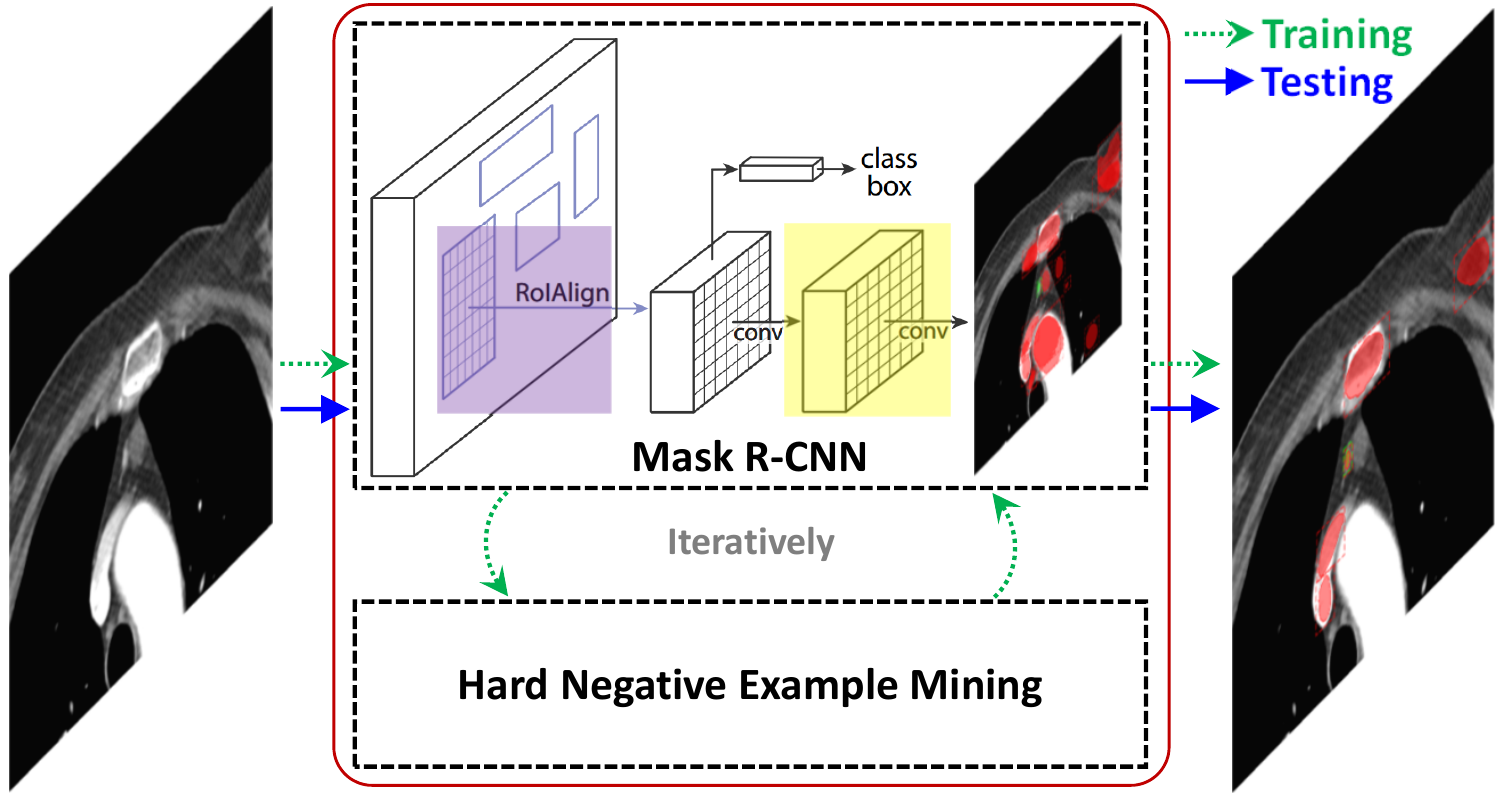}
	\end{center}
	\caption{The framework of the proposed universal lesion detection method (ULDor).}
	\label{fig:framework}
\end{figure}

The training and testing stages of ULDor are shown in Fig. \ref{fig:framework}. For the training stage, the lesion detection results are first obtained using Mask R-CNN \cite{he2017mask}. Then the proposed hard negative example mining (denoted HNEM) strategy is applied to every input. The mined examples are considered as a false positive class, which is combined with the lesion class to re-train the Mask R-CNN model. The above processes are conducted iteratively until meeting the stopping criteria. For the testing stage, given the input CT scans, the lesion detection results are directly produced by the learned Mask R-CNN model with a single inference. Therefore, as shown in Fig. \ref{fig:framework}, there are two main parts in the proposed ULDor, \ie Mask R-CNN and HNEM, which will be elaborated next.

\subsection{Mask R-CNN}
In Section \ref{sec:intro}, we have given a simple description about Faster R-CNN \cite{ren2017faster}, which has two outputs for each RoI, \ie a class label and a bounding-box offset. Extending Faster R-CNN, Mask R-CNN \cite{he2017mask} also outputs a binary mask by adding a branch for predicting segmentation masks (see the yellow region in Fig. \ref{fig:framework}). Experimental results in \cite{he2017mask} demonstrated that constructing the mask branch properly is critical for good results, including the detection ones. But the additional mask output requires extraction of much finer spatial layout of an object. Faster R-CNN is not designed for pixel-to-pixel alignment between inputs and outputs, thus a simple and quantization-free layer (RoIAlign, see the purple region in Fig. \ref{fig:framework}) is proposed in Mask R-CNN to faithfully preserve exact spatial locations. Through experiments in \cite{he2017mask}, RoIAlign leads to large improvements. Inspired by these facts, this paper employs Mask R-CNN to develop a universal lesion detector. Please refer to \cite{he2017mask} for the details of Mask R-CNN.

\begin{figure}[t!]
	\begin{minipage}[b]{1.0\linewidth}
		\centering
		\includegraphics[height=0.32\linewidth]{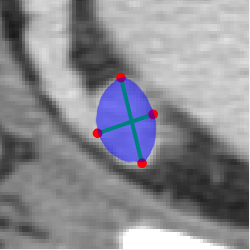}
		\includegraphics[height=0.32\linewidth]{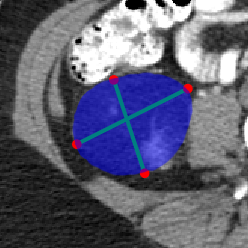}
		\includegraphics[height=0.32\linewidth]{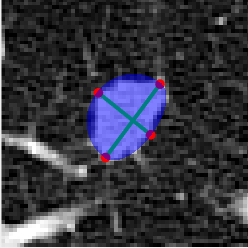}
	\end{minipage}
	
	\begin{minipage}[b]{0.32\linewidth}
		\centering
		\vspace{0.2cm}
		\centerline{(a)}\medskip
	\end{minipage}
	\begin{minipage}[b]{0.32\linewidth}
		\centering
		\vspace{0.2cm}
		\centerline{(b)}\medskip
	\end{minipage}
	\begin{minipage}[b]{0.32\linewidth}
		\centering
		\vspace{0.2cm}
		\centerline{(c)}\medskip
	\end{minipage}
	\hfill
	\caption{Three examples of pseudo mask construction results. The green diameters represent the RECIST annotations, the red spots represent the endpoints of the RECISTs, and the blue areas represent the constructed pseudo masks.}
	\label{fig:psuedo-mask}
\end{figure}

\textbf{Pseudo Mask Construction}. As a large-scale lesion set, the NIH DeepLesion dataset \cite{yan2018deeplesion} only provides the RECIST annotation for each marked lesion region due to high cost of manually annotating lesion masks. But to feasibly and effectively train a Mask R-CNN model, the lesions' masks are required. To address this problem, we construct a pseudo mask for each lesion region from its RECIST annotation that can be considered as a surrogate of the real mask. Fig. \ref{fig:psuedo-mask} shows three examples of pseudo mask construction results. As shown in Fig. \ref{fig:psuedo-mask}, given a lesion region, four endpoints (the red spots) of its RECIST annotation (the green long and short diameters) form four pairs of points from different diameters. An ellipse can be fitted according to each pair of points that are the endpoints of the major and minor axes of the fitted ellipse, whose quarter covering these two points is selected as the constructed mask for this pair. All of these four constructed masks are combined as the pseudo mask of the given lesion region (the blue area).

\subsection{Hard Negative Example Mining}
With the constructed pseudo masks, the Mask R-CNN model can be learned for lesion detection. Through experiments, we find that there are still many false positives in the detection results, which lower the orders of true positives' scores sometimes, so as to harm the detection performance. To alleviate this problem, a hard negative example mining (denoted HNEM) strategy is proposed in this paper.

\begin{figure*}[t!]
	\begin{minipage}[b]{1.0\linewidth}
		\centering
		\includegraphics[height=0.37\linewidth]{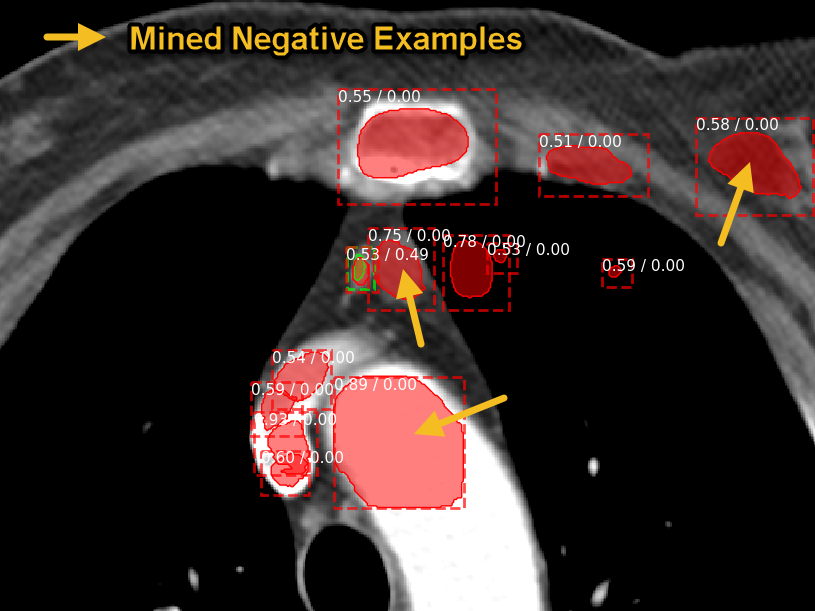}
		\includegraphics[height=0.37\linewidth]{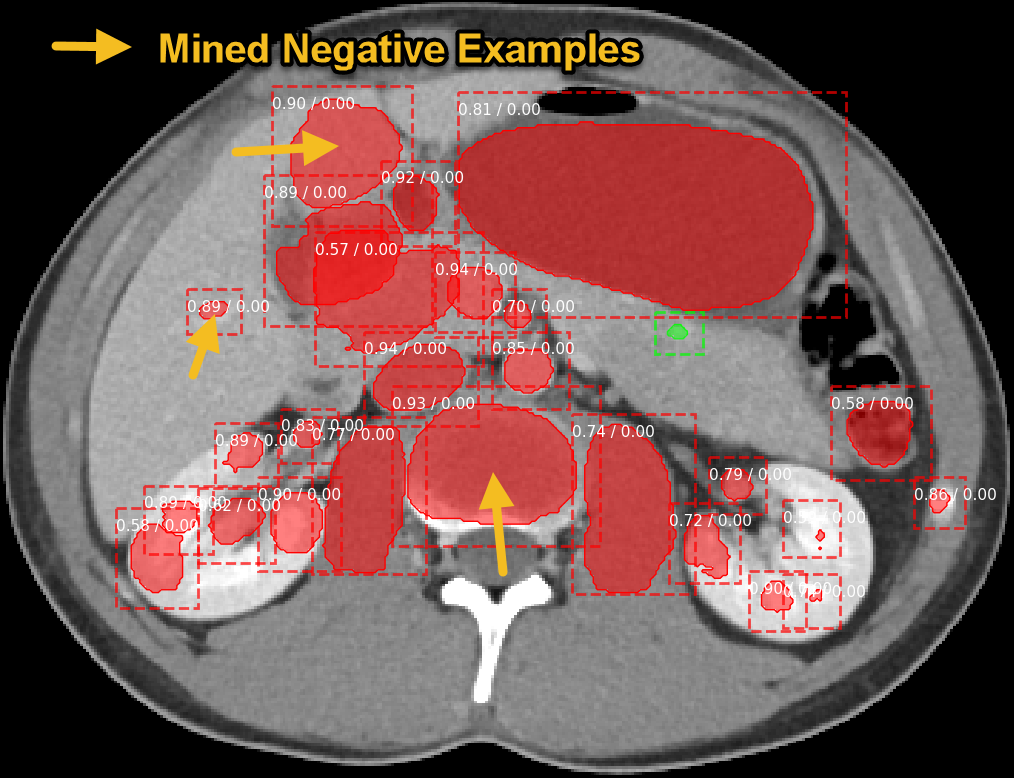}
	\end{minipage}
	\begin{minipage}[b]{0.49\linewidth}
		\centering
		\vspace{0.2cm}
		\centerline{(a)}\medskip
	\end{minipage}
	\begin{minipage}[b]{0.49\linewidth}
		\centering
		\vspace{0.2cm}
		\centerline{(b)}\medskip
	\end{minipage}
	\caption{Two example of HNEM results. For each detected region, a bounding-box, a segmentation mask and a pair of numbers ($S_1/S_2$) are provided. The detection results and reference standard are colored red and green, respectively.}
	\label{fig:HNEM}
\end{figure*}

Given a CT scan, the learned Mask R-CNN produces three outputs, \ie the results of lesion classification, detection and segmentation, from which we can calculate a classification score ($S_1$) and a segmentation overlapping score ($S_2$). Actually, $S_1$ is the probability of a detected region being lesion and $S_2$ is the intersection-over-union (IoU) value between the segmentation result and the pseudo mask. According to $S_1$ and $S_2$, the proposed HNEM strategy proceeds as follows. For an input CT scan at training stage, we first run the previous trained Mask R-CNN model to get $n$ detection results classified as lesion class (denoted \{$D^1, D^2, ..., D^n$\}), and then whose \{$S^1_1, S^2_1, ..., S^n_1$\} and \{$S^1_2, S^2_2, ..., S^n_2$\} are computed. If there exists a detected result $D^i$ meeting $S^i_2>0.3$, it is a reasonably good detection. We randomly choose $ m $ results that have even higher probability, \ie from set $E=\{D^j|S^j_1>S^i_1, 1 \leq j \leq n, j\neq i\}$. Such results are likely negative samples mis-classified as positive ones and with high scores, so we can use them as hard negative examples to re-train and improve our model. If the detection is so poor that none of the detected results is accurate enough, \ie $S^i_2\leq0.3$, for all $ i $, we randomly choose $ m $ results from those with high probability scores, \ie from set $E=\{D^j|S^j_1>0.7, 1 \leq j \leq n\}$. Such results are again negative samples mis-classified with high scores, which we use as hard negative examples.


Here, we set $m=3$ if $\sharp(E)>3$, otherwise, $m=\sharp(E)$, where $\sharp(E)$ is an operation of computing the number of elements in set $E$. Fig. \ref{fig:HNEM} shows two examples of the mined results, where (a) is the case of $S^i_2>0.3$ and $\sharp(E)>3$ and (b) is the case of $S^i_2\leq0.3, 1 \leq i \leq n$ and $\sharp(E)>3$.

\subsection{Implementation Details}
The mined negative examples from entire training set are considered as a false positive class combined with the lesion class to re-train the Mask R-CNN model. The processes of Mask R-CNN training and HNEM are iteratively conducted for every epoch until minimizing the validation loss. In this work, ResNet-101 \cite{he2016deep} is adopted as the backbone of Mask R-CNN. We train all methods using SGD with a weight decay of 0.0001 and momentum of 0.9. The initial learning rate is 0.001 and we decay the learning rate by 0.1 every 4 epochs.

\section{Experiments}
\subsection{Dataset and Evaluation Criterion}
The NIH DeepLesion dataset \cite{yan2018deeplesion} is used for performance evaluation, composed of $32,735$ PACS CT lesion images annotated with RECIST long and short diameters. These are derived from $10,594$ studies of $4,459$ patients. Different from existing specific-type lesion datasets, the DeepLesion dataset contains a variety of lesions across whole body parts including those in lungs, livers, kidneys, etc., and enlarged lymph nodes in the chest, abdomen, and pelvis. Following the split of \cite{yan2018deeplesion}, the DeepLesion dataset is divided into training (70\%), validation (15\%) and test (15\%) sets. A detection result is considered as correct when the IoU between the predicted bounding-box and the real bounding-box is larger than 0.5. The criterion, \ie free-response receiver operating characteristic (FROC) curve is used for evaluation.

\subsection{Results and Analyses}
\begin{figure*}[t!]
	\begin{minipage}[b]{1.0\linewidth}
		\centering
		\includegraphics[trim=200 320 200 315,clip, width=0.32\linewidth]{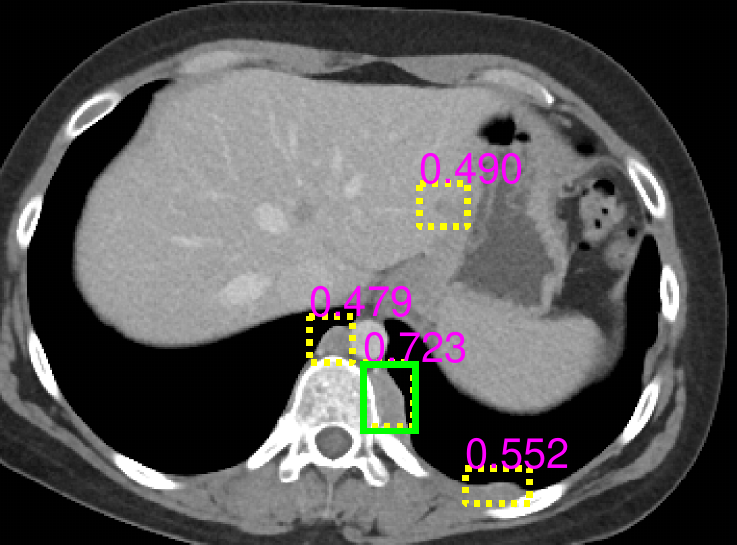}
		\includegraphics[trim=200 320 200 315,clip, width=0.32\linewidth]{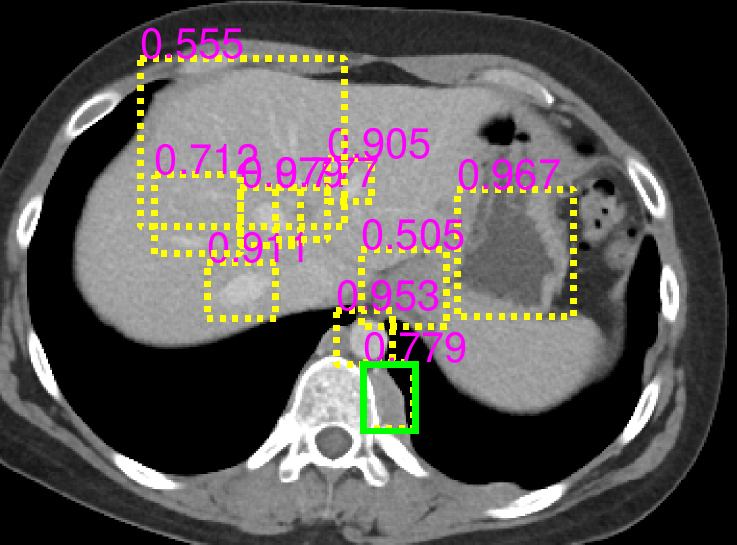}
		\includegraphics[trim=200 320 200 315,clip, width=0.32\linewidth]{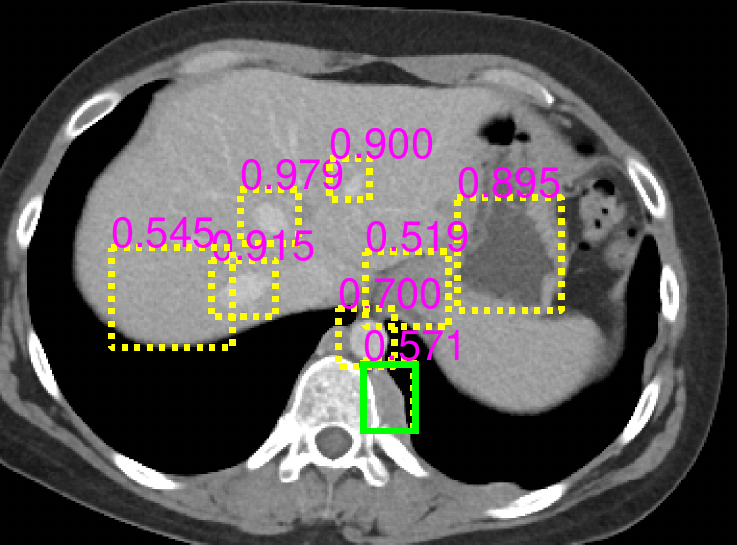} \\
		\vspace{0.05cm}
		\includegraphics[trim=190 325 190 300,clip, width=0.32\linewidth]{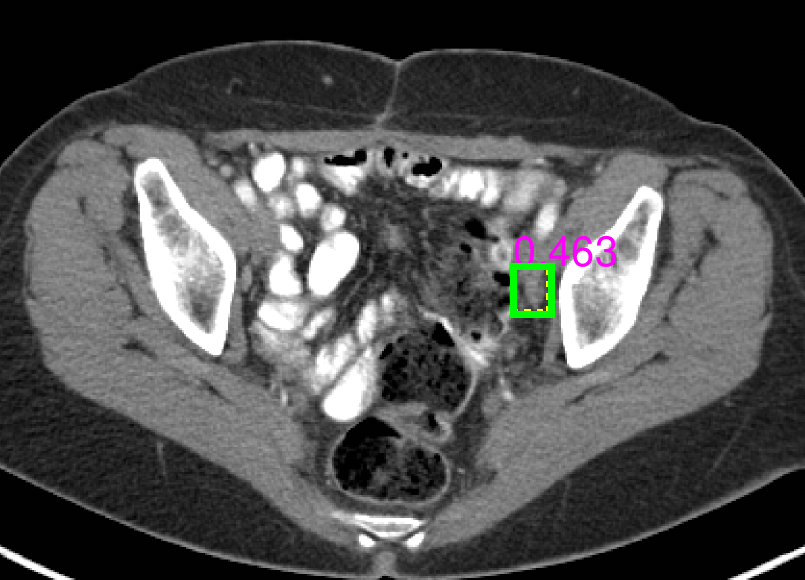}
		\includegraphics[trim=190 325 190 300,clip, width=0.32\linewidth]{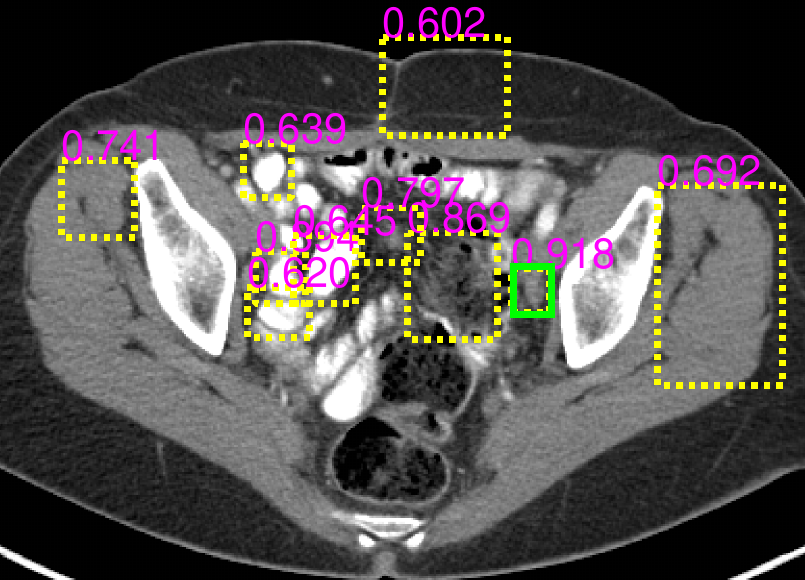}
		\includegraphics[trim=190 325 190 300,clip, width=0.32\linewidth]{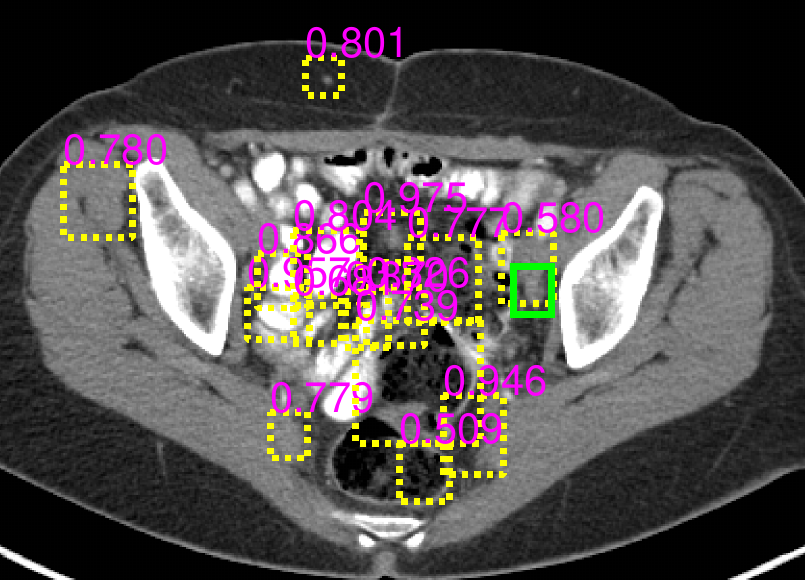} \\
		\vspace{0.05cm}
		\includegraphics[trim=190 325 190 290,clip, width=0.32\linewidth]{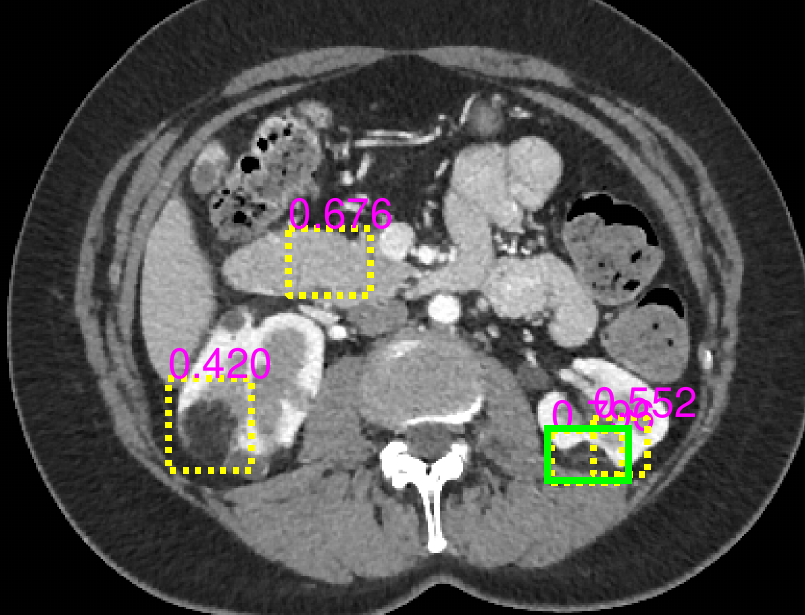}
		\includegraphics[trim=190 325 190 290,clip, width=0.32\linewidth]{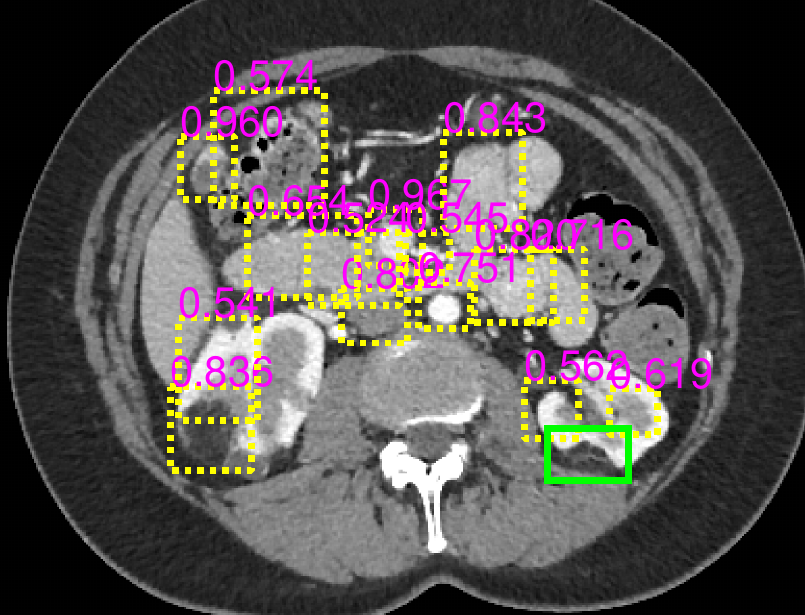}
		\includegraphics[trim=190 325 190 290,clip, width=0.32\linewidth]{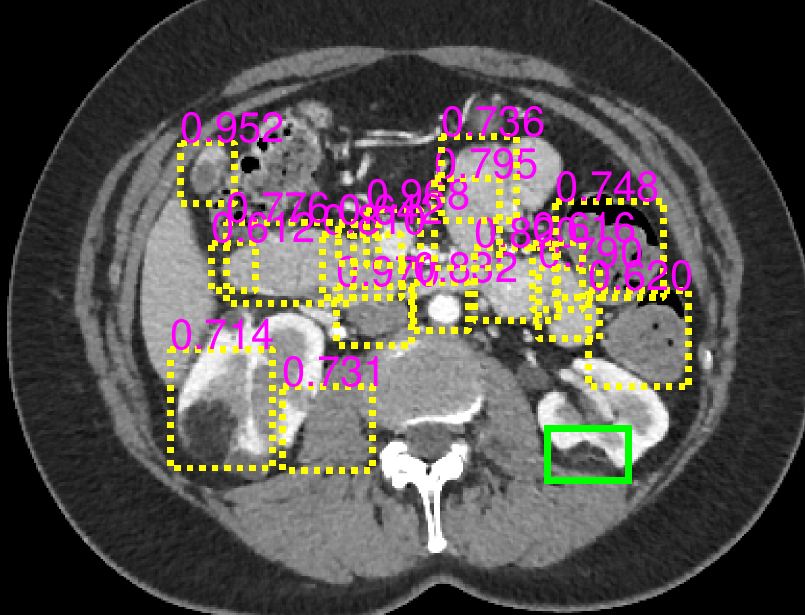} \\
		\vspace{0.05cm}
		\includegraphics[trim=180 330 200 280,clip, width=0.32\linewidth]{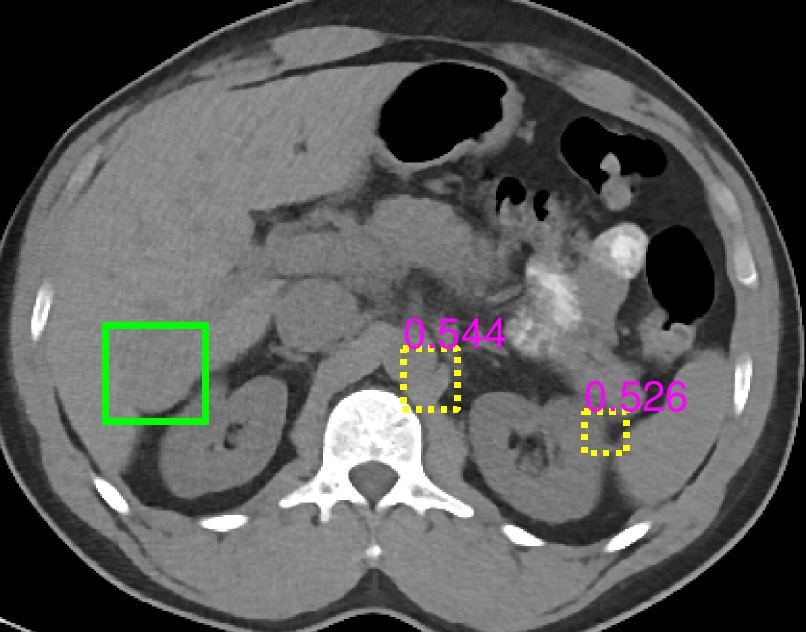}
		\includegraphics[trim=180 330 200 280,clip, width=0.32\linewidth]{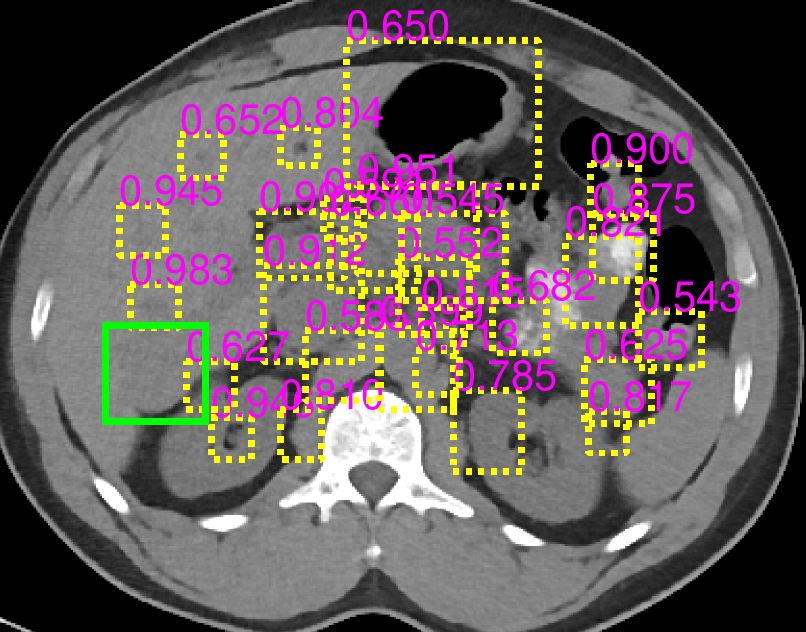}
		\includegraphics[trim=180 330 200 280,clip, width=0.32\linewidth]{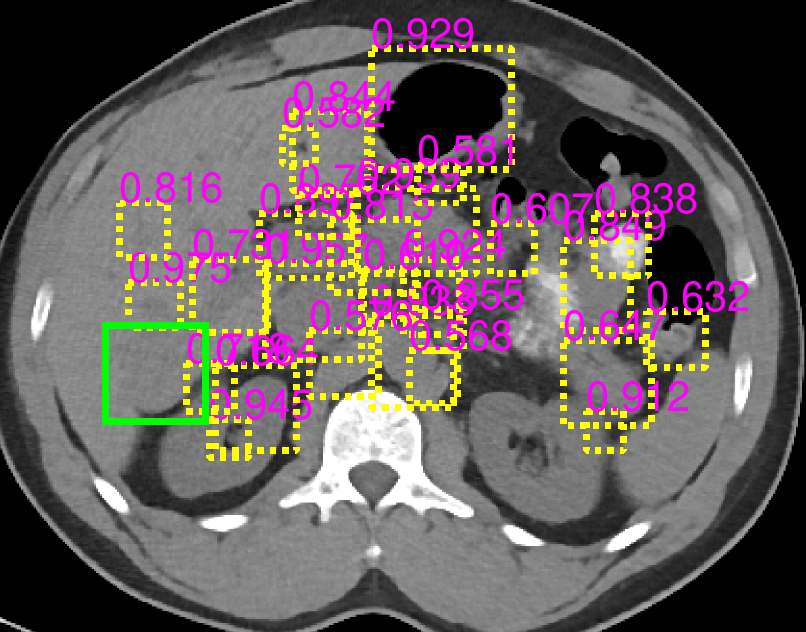} \\
	\end{minipage}
	\begin{minipage}[b]{0.32\linewidth}
		\centering
		\vspace{0.2cm}
		\centerline{(a)}\medskip
	\end{minipage}
	\begin{minipage}[b]{0.32\linewidth}
		\centering
		\vspace{0.2cm}
		\centerline{(b)}\medskip
	\end{minipage}
	\begin{minipage}[b]{0.32\linewidth}
		\centering
		\vspace{0.2cm}
		\centerline{(c)}\medskip
	\end{minipage}
	\caption{Four examples of lesion detection results produced by three methods, \ie (a) Mask R-CNN + HNEM, (b) Mask R-CNN and (c) Mask R-CNN w/o Mask. The green solid bounding-boxes represent the reference standard, the yellow dashed bounding-boxes represent the lesion detection results, and the purple texts represent the classification scores ($S_1$).}
	\label{fig:res}
\end{figure*}

To demonstrate the benefits of using pseudo masks and the hard negative example mining (HNEM) strategy, we conduct the following experimental comparisons. (1) The Mask R-CNN model is trained without using the segmentation branch and HNEM (denoted Mask R-CNN w/o Mask). (2) The Mask R-CNN model is trained with pseudo masks but without HNEM (denoted Mask R-CNN). (3) The Mask R-CNN model is trained with pseudo masks and HNEM (denoted Mask R-CNN + HNEM).

\textbf{Qualitative Comparisons}. Fig. \ref{fig:res} shows four qualitative lesion detection results produced by the models trained with above configurations. From Fig. \ref{fig:res}, we can see that (1) the correct detection results in Fig. \ref{fig:res}(b) get higher classification scores than the ones in Fig. \ref{fig:res}(c) and the true lesion in the second row can be correctly detected in Fig. \ref{fig:res}(b) while failed in Fig. \ref{fig:res}(c), meaning that the Mask R-CNN model can be learned more effectively with pseudo masks to improve the detection performance. (2) The false positives in Fig. \ref{fig:res}(a) are much less than the ones in Fig. \ref{fig:res}(b), the ranking order of true positive in the first row of Fig. \ref{fig:res}(a) is more forward than the one in Fig. \ref{fig:res}(b), and the true lesion in the third row can be correctly detected in Fig. \ref{fig:res}(a) while failed in Fig. \ref{fig:res}(b), meaning that with the help of mined negative examples, the Mask R-CNN model can learn more discriminative features to distinguish true lesions from false positives, so as to reduce the false positives and improve the detection performance. These intuitively demonstrate that the pseudo masks and the hard negative example mining strategy are useful for lesion detection.
Additionally, the last row of Fig. \ref{fig:res} gives a failure example of lesion detection. As we can see that the proposed ULDor cannot well detect the lesions when they and their surrounding tissues have very similar appearances.

\begin{figure}[t!]
	\begin{center}
		\includegraphics[width=0.99\linewidth]{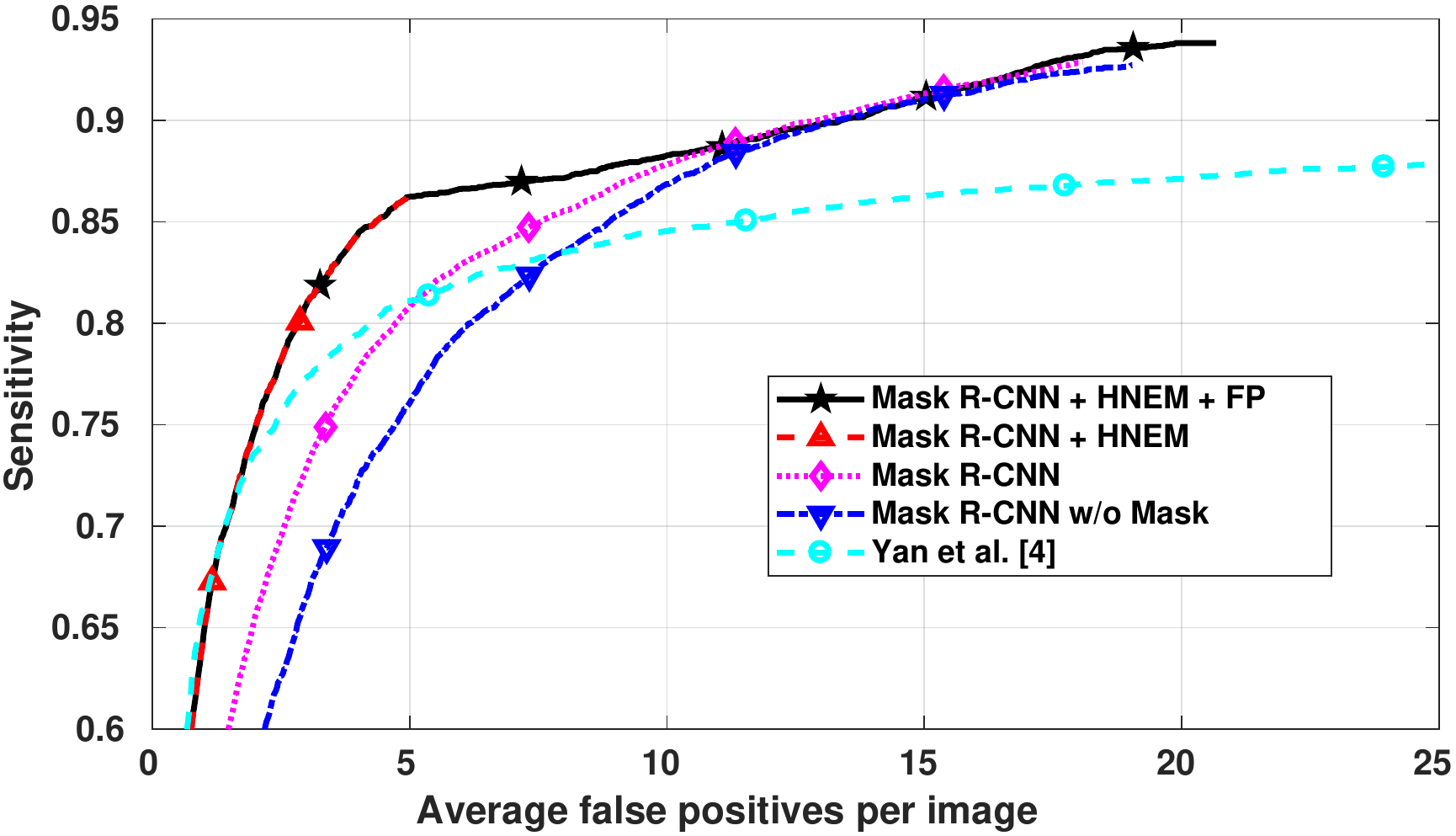}
	\end{center}
	\caption{FROC curves of various methods on the test set.}
	\label{fig:FROC}
\end{figure}

\begin{table}[t!]
	\begin{center}
		\caption{Sensitivity (\%) at several specific average false positives per image on FROC curves.}
		{
			\scriptsize
			
			\begin{tabular}{|@{}*{1}{m{3.32cm}<{\centering}@{}|@{}}*{6}{m{0.85cm}<{\centering}@{}|@{}}}
				\hline
				Method & 0.5 & 1 & 2 & 4 & 8 & 16 \\ \hline
				\hline
				Yan et al. \cite{yan2018deeplesion}  &  \textbf{55.51} & \textbf{66.13}  &  73.64 & 79.44  &  83.52 & 86.50\\ \hline
				\textbf{Mask R-CNN w/o Mask}  &  32.51 &  44.90 & 58.61  & 72.30  & 83.56  & 91.46\\ \hline
				\textbf{Mask R-CNN}  & 39.82  & 52.66  & 65.58  & 77.73  & 85.54  & \textbf{91.80}\\ \hline
				\textbf{Mask R-CNN + HNEM}  &   52.86  & 64.80  & \textbf{74.84}  & \textbf{84.38}   & -  & -\\ \hline 
				\textbf{Mask R-CNN + HNEM + FP}     & 52.86 & 64.80  & \textbf{74.84}  & \textbf{84.38}  & \textbf{87.17}  & \textbf{91.80}\\\hline
			\end{tabular}
		}
	\end{center}
	\label{tab:results}
\end{table}

\textbf{Quantitative Comparisons}. The quantitative results are displayed in Fig. \ref{fig:FROC} and Table 1, including those reported in \cite{yan2018deeplesion}. As we can see that (1) Mask R-CNN always gets higher sensitivity than Mask R-CNN w/o Mask, suggesting the power of Mask R-CNN is enhanced with pseudo masks for lesion detection. (2) Mask R-CNN + HNEM always gets higher sensitivity than Mask R-CNN with 5 or less false positives (FPs) per image, suggesting HNEM is helpful to make Mask R-CNN learn more discriminative features for performance improvement. (3) The maximum of average false positives (AFP) per image is 5 in the detection results produced by Mask R-CNN + HNEM, meaning that the FPs are reduced effectively using HNEM. (4) As AFP increasing, the largest sensitivity of Mask R-CNN + HNEM is 86.21\% that is smaller than the other methods. That's because there are some true lesions that are the missing annotations in the mined negative examples. After training Mask R-CNN with these mined samples, the model will classify some true lesions that are similar to the mined samples as FPs during testing. To verify this point, we consider more and more FPs from the FP class as the detection results, then we compute the quantitative results (see the results of Mask R-CNN + HNEM + FP in Fig. \ref{fig:FROC} and Table 1), which are better than the ones of other methods as AFP increasing, meaning that some true lesions are really classified as FP class. The largest sensitivity is about 94.2\%, while the one in \cite{yan2018deeplesion} is about 88.0\%. Although we gain a large improvement, we still miss about 5.8\% of the true lesions. These missed lesions may cause high risk for patients in real clinical diagnosis. Therefore, there is still a large improvement space we can pursue. (5) Compared with \cite{yan2018deeplesion}, our method gets worse performance when APF is less than 1. The possible reason is that VGGNet-16 \cite{simonyan2014very} is adopted as the backbone in \cite{yan2018deeplesion} that has more parameters (138M) than ResNet-101 (44.5M) used in this work, leading to learn more representative features to detect more top-1 true positives.

\section{Conclusions}
We propose a universal lesion detection method based on Mask R-CNN to detect all kinds of lesions from CT scans. In the proposed method, pseudo masks of lesion regions are constructed to effectively train the Mask R-CNN model and a hard negative example mining (HNEM) strategy is proposed for false positive reduction. The processes of Mask R-CNN training and hard negative example mining are conducted iteratively to get the final model for lesion detection. Experimental results on the NIH DeepLesion dataset demonstrate that the lesion detection performance can be boosted using pseudo masks and HNEM. The proposed method achieves a sensitivity of 86.21\% with five false positives per image, which obtains an absolute improvement of 5.11\% from 81.1\% reported in \cite{yan2018deeplesion}, suggesting that the proposed method potentially provides a highly positive impact to clinical workflows, \eg coupling it with some tools (\eg lesion segmentation) to make them fully automatic. 

\vspace*{0.5\baselineskip}
\noindent\textbf{Acknowledgments.}
This research was supported by the Intramural Research Program of the National Institutes of Health Clinical Center and by the Ping An Insurance Company through a Cooperative Research and Development Agreement. We thank Nvidia for GPU card donation.

\bibliographystyle{IEEEbib}
\bibliography{refs}

\end{document}